\documentclass{article}
\usepackage{ijcai26}

\usepackage{times}
\usepackage{soul}
\usepackage{url}
\usepackage[hidelinks]{hyperref}
\usepackage[utf8]{inputenc}
\usepackage[small]{caption}
\usepackage{graphicx}
\usepackage{amsmath}
\usepackage{amsthm}
\usepackage{booktabs}
\usepackage{algorithm}
\usepackage{algorithmic}
\usepackage{amssymb}      
\usepackage{amsfonts}
\usepackage{ijcai26}      
\usepackage[switch]{lineno}
\usepackage{xcolor}

\title{Weakly Supervised Spatio-Temporal Candidate Discovery of Dairy Farm Sites from Seasonal Satellite Imagery}

\author{
Usman Haider$^1$
\and
Fatima Khalid$^2$\and
Karl Mason$^{1}$
\affiliations
$^1$School of Computer Science, University of Galway, Ireland.\\
$^2$Faculty of Computer Science and Engineering, GIK Institute of Engineering Sciences and Technology, Pakistan.\\
\emails
\{usman.haider, karl.mason\}@universityofgalway.ie
}

\begin{document}

\maketitle
\begin{abstract}
{\color{black}
Farm site discovery from satellite imagery is a spatiotemporal candidate ranking problem because farm evidence is distributed across pasture, field boundaries, roads, buildings, and seasonal vegetation patterns. Direct farm labels are often incomplete, which makes fully supervised detection difficult. This paper proposes a weakly supervised pipeline for ranking dairy farm candidate clusters from seasonal Sentinel imagery and open map priors. The method uses aligned spring, summer, and autumn image tiles from County Cork, Ireland, with spectral bands, vegetation indices, built area indices, and a pasture channel. A Barlow Twins encoder learns multi-season tile embeddings without farm labels. In parallel, weak OpenStreetMap farm priors are split into a prior and a held-out set. Prior features support a rule-based tile score that combines farm proximity, seasonal pasture evidence, and summer greenness, while held-out features are reserved only for proxy evaluation. The rule score is smoothed over a spatial representation graph using geographic proximity and embedding similarity, and high-scoring tiles are grouped into ranked candidate clusters. From 26,722 valid tiles, the main run selects 535 high-confidence tiles and forms 71 candidate clusters. The top 5 clusters achieve 0.60 precision within 500 m and 0.80 precision within 1000 m of held-out OpenStreetMap farm features. The top 10 clusters achieve 0.40 precision within 500 m and 0.80 precision within 1000 m. The results show that seasonal representation learning and weak geographic priors can reduce large satellite image collections into compact candidate sets for human review.
}
\end{abstract}

\section{Introduction}
\label{sec:introduction}

Discovering farms from satellite imagery is a challenging, weakly supervised spatiotemporal reasoning problem. Farms are rarely localized as isolated visual objects. Instead, it appears as a spatial configuration of pasture, field boundaries, roads, buildings, yards, vegetation state, and seasonal management patterns. These cues may span hundreds of meters and can resemble villages, rural housing, industrial sites, forests, or other agricultural land uses. Remote sensing benchmarks have shown the value of deep learning for land use, land cover, and scene understanding, but they often assume explicit labels or dense annotations \cite{helber2019eurosat,demir2018deepglobe}. Farm candidate discovery is harder because farm locations are incompletely mapped, farm boundaries are often unavailable, and labels distinguishing farm types are sparse or noisy.

This paper studies weak supervision for discovering dairy farm candidates from seasonal satellite imagery. Rather than producing a definitive farm inventory, the goal is to rank spatial clusters that are likely to correspond to farm-associated sites. This setting is useful where direct farm labels are incomplete, but open map data provides weak geographic cues. OpenStreetMap has been used to support remote sensing and land use mapping, although its coverage and tagging quality vary spatially \cite{estima2015osm,johnson2022opensentinelmap}. Open geographic data can provide weak labels, priors, and contextual signals for land-use analysis \cite{estima2013osm,li2022leveraging,schultz2025osmlanduse}. These data are therefore useful as weak supervision, but not as definitive ground truth.

The proposed framework combines seasonal remote sensing cues, weak geographic priors, self-supervised representation learning, and graph-based spatial reasoning. We use aligned spring, summer, and autumn multispectral imagery from County Cork, Ireland, a pasture-rich region with rural settlements, farm complexes, coastal areas, and urban centers. Each tile contains Sentinel spectral bands, vegetation indices, water and built area indices, and a pasture channel. We train a Barlow Twins self-supervised encoder on multi-season tiles to learn visual representations without farm labels \cite{chen2020simclr,zbontar2021barlow}. Recent work has shown that self-supervised learning can exploit temporal and spectral structure in satellite imagery \cite{manas2021seasonal,cong2022satmae}. Still, here the learned embeddings are used for graph smoothing and candidate ranking rather than supervised classification.

{\color{black}
Spatial context is central to the task. A single tile may contain pasture, while nearby tiles contain roads, field boundaries, or related agricultural structures. Our pipeline first assigns each tile a rule-based candidate score from seasonal pasture, summer greenness, and weak OpenStreetMap farm proximity. The score is then refined via graph-based weak-label propagation using geographic proximity and self-supervised embedding similarity. High-scoring tiles are grouped by grid adjacency to form ranked candidate clusters.
}
\begin{figure*}[!htbp]
    \centering
    \includegraphics[width=0.95\linewidth]{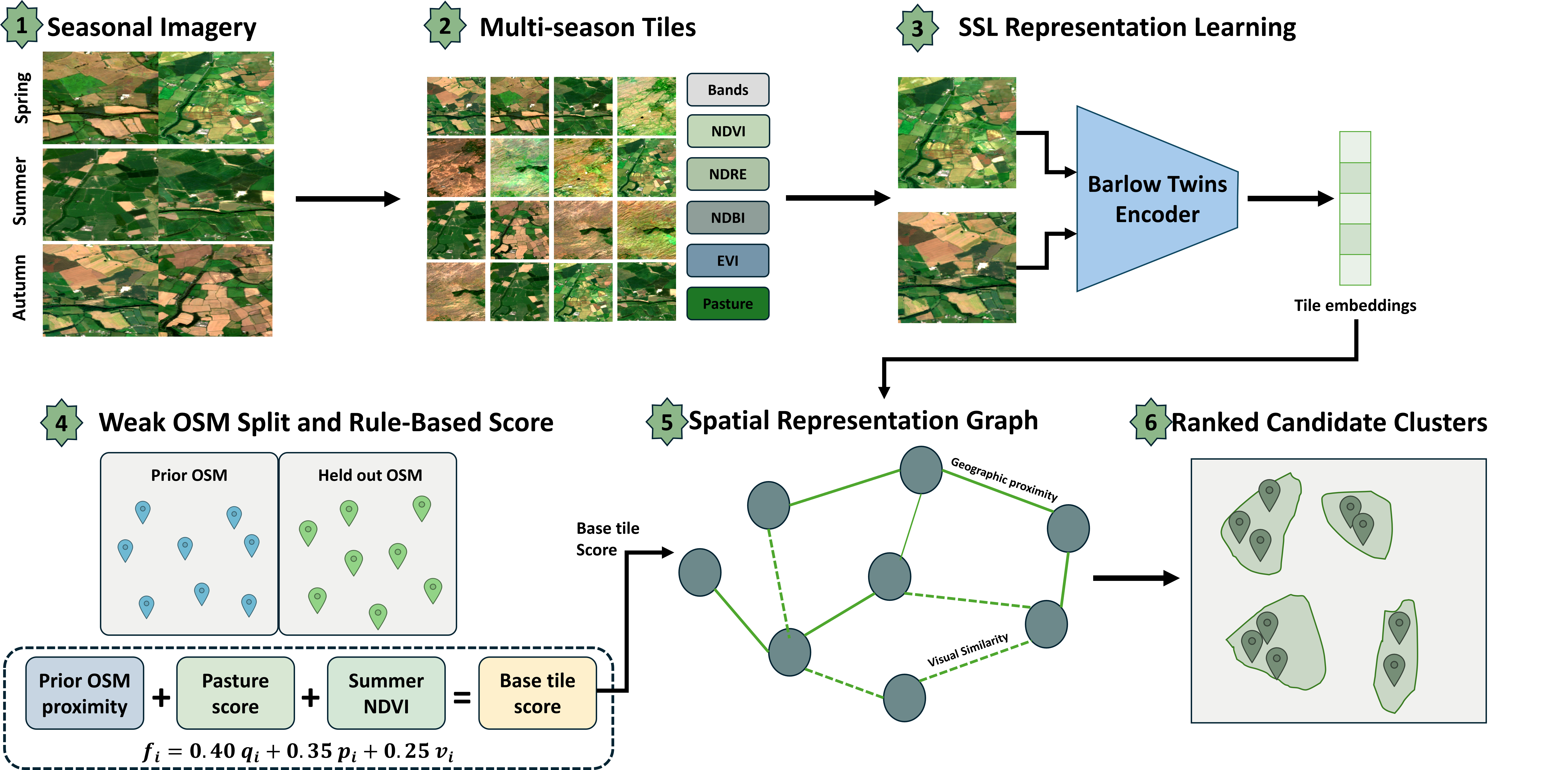}
    \caption{Weakly supervised farm candidate discovery pipeline from seasonal imagery, showing multi-season tiling, SSL embedding extraction, OSM prior-based scoring, graph smoothing, and ranked candidate clusters.}
    \label{fig:framework}
\end{figure*}
We evaluate candidates using a proxy protocol in which OpenStreetMap farm features are split into a prior and a held-out set. The prior set provides weak farm-proximity cues during scoring, while the held-out set is used solely for distance-based evaluation. Because OpenStreetMap farm features are incomplete and spatially noisy, held-out proximity is not treated as true farm detection accuracy. Instead, it measures whether ranked candidates align with farm-related map features that were not used as farm proximity priors. This follows weak supervision settings in remote sensing, where labels may be noisy, indirect, or derived from external geographic products \cite{wang2020weakly}.
{\color{black}
The main run reduces 26,722 valid multi-season tiles to 535 high-confidence tiles and 71 candidate clusters. When ranked by the smoothed method score, the top 5 clusters achieve 0.60 precision within 500 m and 0.80 precision within 1000 m of held-out OpenStreetMap farm features. The top 10 clusters achieve 0.40 precision within 500 m and 0.80 precision within 1000 m. Across all clusters, 19 are within 500 m, and 46 are within 1000 m of a held-out feature, with a median distance of 744.3 m. Ablations show that NDVI alone, pasture alone, summer-only features, and OSM priors alone are insufficient. The strongest signals in the primary framework are seasonal pasture dynamics, vegetation cues, weak farm proximity, self-supervised representations, and graph-based spatial smoothing.
}
{\color{black}
This paper makes three contributions. First, it presents a weak supervision pipeline for dairy farm candidate discovery using seasonal satellite imagery and open map priors. Second, it combines seasonal pasture, summer greenness, weak farm proximity, self-supervised tile embeddings, graph smoothing, and spatial clustering. Third, it provides a proxy evaluation protocol that separates OpenStreetMap farm features into prior and held-out sets, while making clear that the resulting metrics measure candidate ranking against noisy map evidence rather than definitive farm detection accuracy.
}

\section{Methodology}
\label{sec:methodology}

Figure~\ref{fig:framework} summarizes the pipeline. The method takes aligned spring, summer, and autumn satellite imagery and weak OpenStreetMap farm features as input. It extracts multi-season tiles, learns self-supervised tile embeddings, computes rule-based dairy farm candidate scores, smooths scores over a spatial and representation graph, groups high-scoring tiles into clusters, and evaluates ranked clusters against held-out OpenStreetMap farm features.

\subsection{Weak Supervision Setup}
\label{subsec:task}

Let $\mathcal{S}=\{\mathrm{spring},\mathrm{summer},\mathrm{autumn}\}$ be the set of seasons. For each season, we assume spatially aligned satellite imagery over the same region. A tile $i$ is represented by a multi-season tensor
\[
    x_i \in \mathbb{R}^{|\mathcal{S}| \times B \times H \times W},
\]
and a geographic center coordinate $p_i=(\lambda_i,\phi_i)$.

Direct farm labels are not assumed. Instead, OpenStreetMap farm-related features are split into two disjoint sets:
\[
    \mathcal{O}
    =
    \mathcal{O}_{\mathrm{prior}}
    \cup
    \mathcal{O}_{\mathrm{heldout}},
    \quad
    \mathcal{O}_{\mathrm{prior}}
    \cap
    \mathcal{O}_{\mathrm{heldout}}
    =
    \emptyset
\]
{\color{black}
The prior set provides weak farm proximity cues during scoring. The held-out set is used only for proxy evaluation. 
}
\subsection{Seasonal Tiles and Self-Supervised Embeddings}
\label{subsec:tiles_embeddings}

Tiles are extracted at matching grid positions across all seasons. In our experiments, each tile is $128 \times 128$ pixels and has a stride of 64 pixels. For each season, we use six Sentinel spectral bands and six derived channels:
\[
\begin{aligned}
\{&B2,B3,B4,B8,B11,B12, \mathrm{NDVI},\mathrm{NDWI},\\
 &\mathrm{NDBI},\mathrm{NDRE},\mathrm{EVI},\mathrm{pasture}\}.
\end{aligned}
\]
With three seasons, each tile contains 36 channels. Reflectance bands are scaled to $[0,1]$, spectral indices are clipped to $[-1,1]$ and mapped to $[0,1]$, and the pasture channel is clipped to $[0,1]$. Tiles with valid reflectance coverage below 85 percent are removed.

{\color{black}
For each tile, we also compute summary features for diagnostics and ablations, including seasonal pasture ratio, seasonal NDVI statistics, edge magnitude, and a texture measure based on the standard deviation of $B8 - B4$. The primary score uses pasture evidence, summer greenness, and weak farm proximity, while the self-supervised embeddings are used in the graph smoothing stage described below.
}

A ResNet-18 encoder is modified to accept a 36-channel input. It is trained with the Barlow Twins objective using two augmented views of each tile. Augmentations include flips, gain and bias perturbations, noise, and crop resizing. After training, the projection head is discarded, and the encoder output $h_i$ is used as the tile embedding.

\subsection{Farm Candidate Score}
{\color{black}
Each tile receives a rule-based candidate score. The score is not a calibrated probability. It is a ranking signal that combines three evidence streams: weak proximity to prior OpenStreetMap farm features, pasture evidence, and summer greenness.

For tile $i$, let $q_i$ be the normalized proximity to the nearest feature in $O_{\mathrm{prior}}$, let $p_i$ be the pasture-based score, and let $v_i$ be the normalized summer NDVI. The primary rule score is
}

{\color{black}
\[
f_i = 0.40 q_i + 0.35 p_i + 0.25 v_i .
\]
}

{\color{black}
The pasture-based score summarizes pasture evidence across spring, summer, and autumn. Summer greenness is represented separately by normalized summer NDVI. This formulation favors pasture-rich agricultural regions with strong seasonal vegetation evidence and proximity to weak farm priors.
}

\subsection{Graph Smoothing and Clustering}
\label{subsec:graph_clustering}

Scores are smoothed over a spatial-representation graph to incorporate local spatial context and embedding similarity. Each tile is connected to its $k=8$ nearest geographic neighbors using projected coordinates. For tile $i$ and neighbor $j$, the geographic kernel is
\[
    g_{ij}
    =
    \exp
    \left(
    -\frac{\delta_{ij}}{1000}
    \right),
\]
where $\delta_{ij}$ is projected distance in meters. Representation similarity is
\[
    s_{ij}=\max(0,\cos(h_i,h_j)).
\]
The smoothed score is
\[
    f_i^G
    =
    \frac{
        \sum_{j \in \mathcal{N}(i)}
        \left(
            0.20 g_{ij}
            +
            0.40 s_{ij}
        \right)
        f_j
    }{
        \sum_{j \in \mathcal{N}(i)}
        \left(
            0.20 g_{ij}
            +
            0.40 s_{ij}
        \right)
        +
        \epsilon
    }.
\]
The final score is
\[
    \tilde{f}_i
    =
    0.85 f_i
    +
    0.15 f_i^G .
\]
{\color{black}
Tiles above the $92^{nd}$ percentile of $\tilde f_i$ are selected and grouped with eight neighbor grid adjacency using the tile stride as the adjacency step. Each connected component becomes one candidate cluster. The representative coordinate is selected from the tile with the highest smoothed method score in the cluster. Clusters are ranked using representative tile score, mean cluster score, and cluster compactness. For ablations, each score variant uses the same thresholding and clustering procedure, but clusters are ranked by the corresponding method score for direct comparison.
}

\subsection{Proxy Evaluation}
\label{subsec:proxy_eval}

Since complete farm ground truth is unavailable, evaluation uses held-out OpenStreetMap farm features as proxy labels. For each predicted cluster, we compute the distance from its representative coordinate to the nearest feature in $\mathcal{O}_{\mathrm{heldout}}$. A cluster is counted as a proxy match at radius $r$ if it lies within $r$ meters of a held-out feature.

We report precision at 250 m, 500 m, and 1000 m, precision at rank, median distance to held-out features, and held-out feature recall. These metrics measure alignment with withheld map features, not exact farm boundary accuracy or true farm detection accuracy.

\section{Experimental Setup and Results}
\label{sec:experiments_results}

\subsection{Study Area, Data, and Metrics}
\label{subsec:data_metrics}

The proposed framework is evaluated in County Cork, Ireland, a pasture-rich region with rural settlements, farm complexes, coastal areas, towns, and heterogeneous land cover. The satellite input consists of aligned spring, summer, and autumn imagery. Each season contains six Sentinel spectral bands and six derived channels: $B2$, $B3$, $B4$, $B8$, $B11$, $B12$, NDVI, NDWI, NDBI, NDRE, EVI, and a pasture channel. Tiles are extracted at $128 \times 128$ pixels with a stride of 64 pixels. After validity filtering, the final dataset contains 26,722 multi-season tiles.

OpenStreetMap farm-related features are used as weak geographic supervision. Features tagged as farms, farmyards, barns, agricultural buildings, and farm buildings are split into two equal parts. The prior set contains 2,117 features and is used for farm proximity scoring. The held-out set contains 2,117 features and is used only for proxy evaluation. {\color{black} General OSM building features are not used in the primary score. A building aware score is evaluated only as an ablation.}
{\color{black}
Each predicted cluster is represented by the coordinate selected by the clustering procedure. In the main pipeline, this coordinate is taken from the tile with the highest smoothed method score in the cluster.} We measure distance from this coordinate to the nearest held-out OSM farm feature. A cluster is counted as a proxy match at radius $r$ if it lies within $r$ meters of a held-out feature. We report precision at 250 m, 500 m, and 1000 m, as well as precision at rank, median distance, and held-out recall. These are proxy metrics, not definitive farm detection accuracy, because no manually verified farm ground truth is used.

\subsection{Main Candidate Discovery Results}
\label{subsec:main_results}

\begin{figure}[t]
    \centering
    \includegraphics[width=0.92\linewidth]{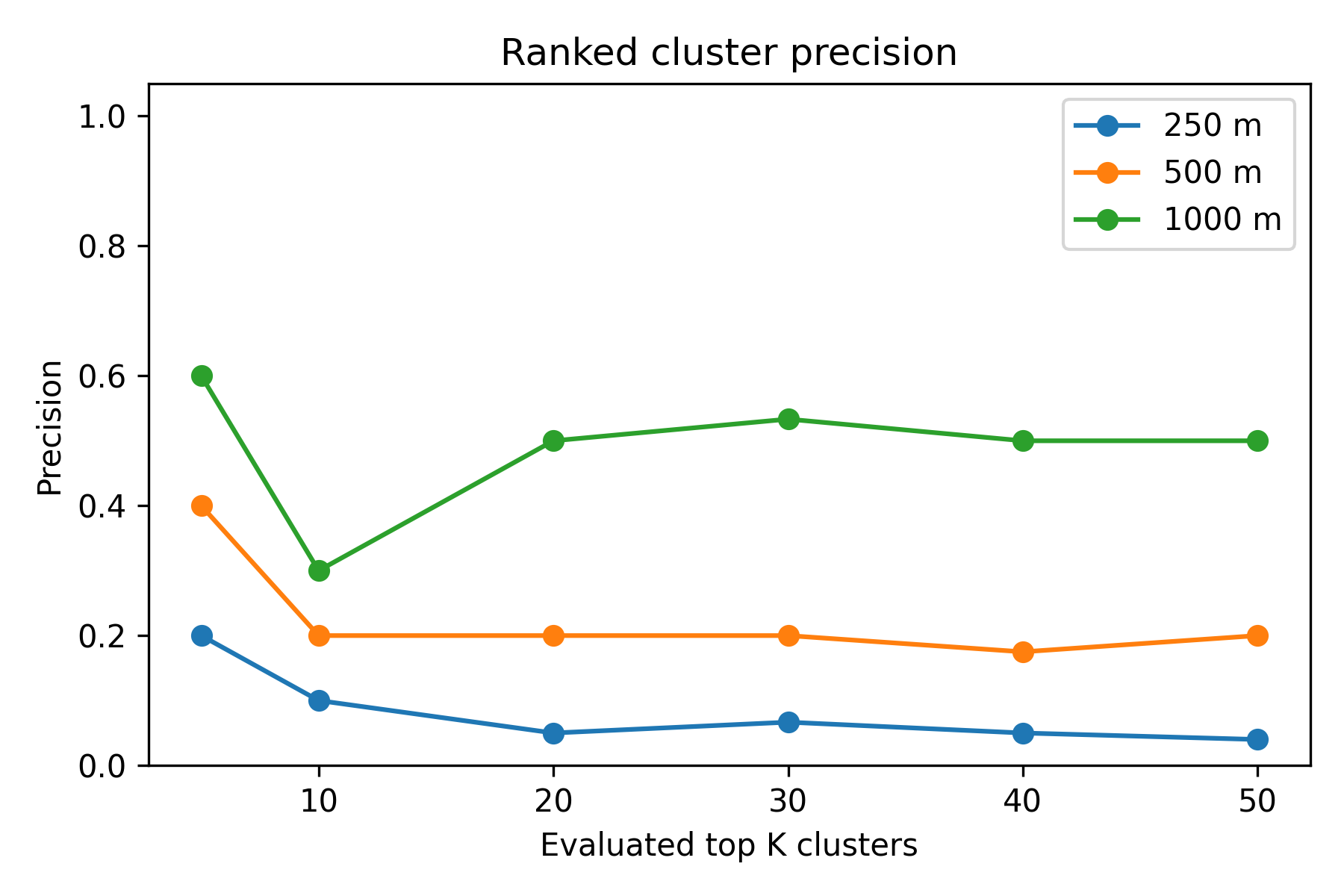}
   \caption{Ranked cluster precision at 250 m, 500 m, and 1000 m. The highest ranked candidates align most strongly with held out OSM farm features, especially at 1000m}

    \label{fig:precision_at_k}
\end{figure}

{\color{black}
The main run selects 535 high-confidence tiles from 26,722 valid tiles and groups them into 71 candidate clusters. Table 1 summarizes the held-out OSM proxy evaluation.

\begin{table}[t]
\centering
{\color{black}
\caption{Proxy validation for the main candidate discovery run.}
\label{tab:main_results}
\begin{tabular}{lr}
\hline
Metric & Value \\
\hline
Total tiles & 26,722 \\
High confidence tiles & 535 \\
Candidate clusters & 71 \\
Held out OSM farm features & 2,117 \\
Median distance to held out OSM farm & 744.3 m \\
Clusters within 250 m & 10 \\
Clusters within 500 m & 19 \\
Clusters within 1000 m & 46 \\
Top 5 precision at 500 m & 0.60 \\
Top 10 precision at 500 m & 0.40 \\
Top 10 precision at 1000 m & 0.80 \\
Held out recall at 1000 m & 0.047 \\
\hline
\end{tabular}
}
\end{table}

The framework reduces a large multi-season tile collection into a compact ranked set of high-confidence farm candidates. Since only 71 clusters are predicted for 2,117 held-out OSM features, the recall is low. This behavior is consistent with the candidate-discovery objective of the framework, not exhaustive inventory mapping. Precision at rank is therefore the more relevant metric. The top 5 candidates reach 0.60 precision within 500 m and 0.80 precision within 1000 m. The top 10 candidates reach 0.40 precision within 500 m and 0.80 precision within 1000 m. Across all predicted clusters, 19 of 71 are within 500 m, and 46 of 71 are within 1000 m of a held-out OSM farm feature.

Figure~\ref{fig:precision_at_k} shows that the candidate list is strongest near the top of the ranking. At the top 5, precision reaches 0.20 within 250 m, 0.60 within 500 m, and 0.80 within 1000 m. At rank 50, 66\% of candidates fall within 1000\,m of a held-out farm feature. This supports using the method as a candidate ranking tool rather than a complete farm map.
}

\subsection{Spatial and Qualitative Evidence}
\label{subsec:spatial_qualitative}
\begin{figure*}[t]
    \centering
    \begin{minipage}{0.49\linewidth}
        \centering
        \includegraphics[width=1.1\linewidth]{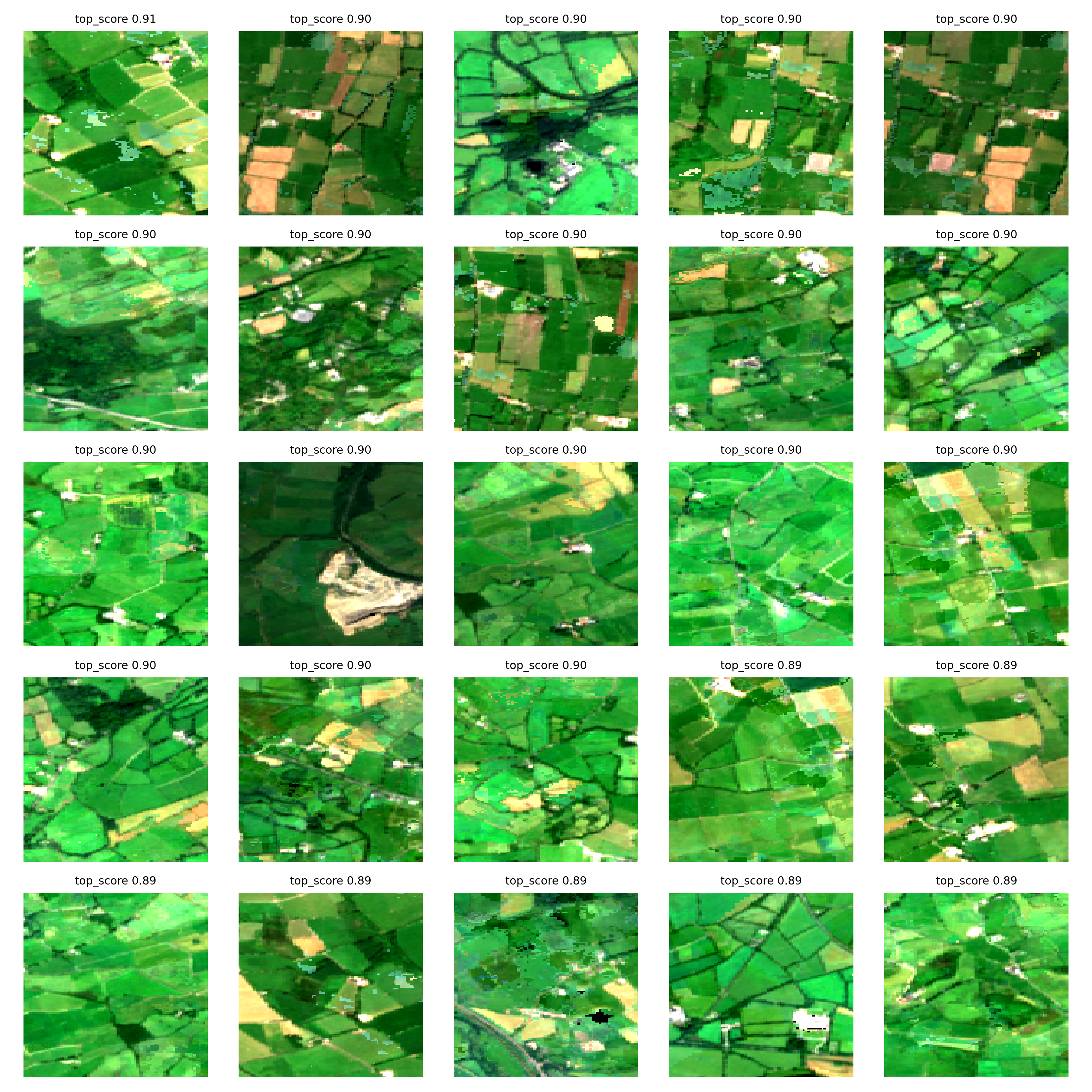}
        \centerline{Top score}
    \end{minipage}
    \hfill
    \hfill
    \begin{minipage}{0.49\linewidth}
        \centering
        \includegraphics[width=1.1\linewidth]{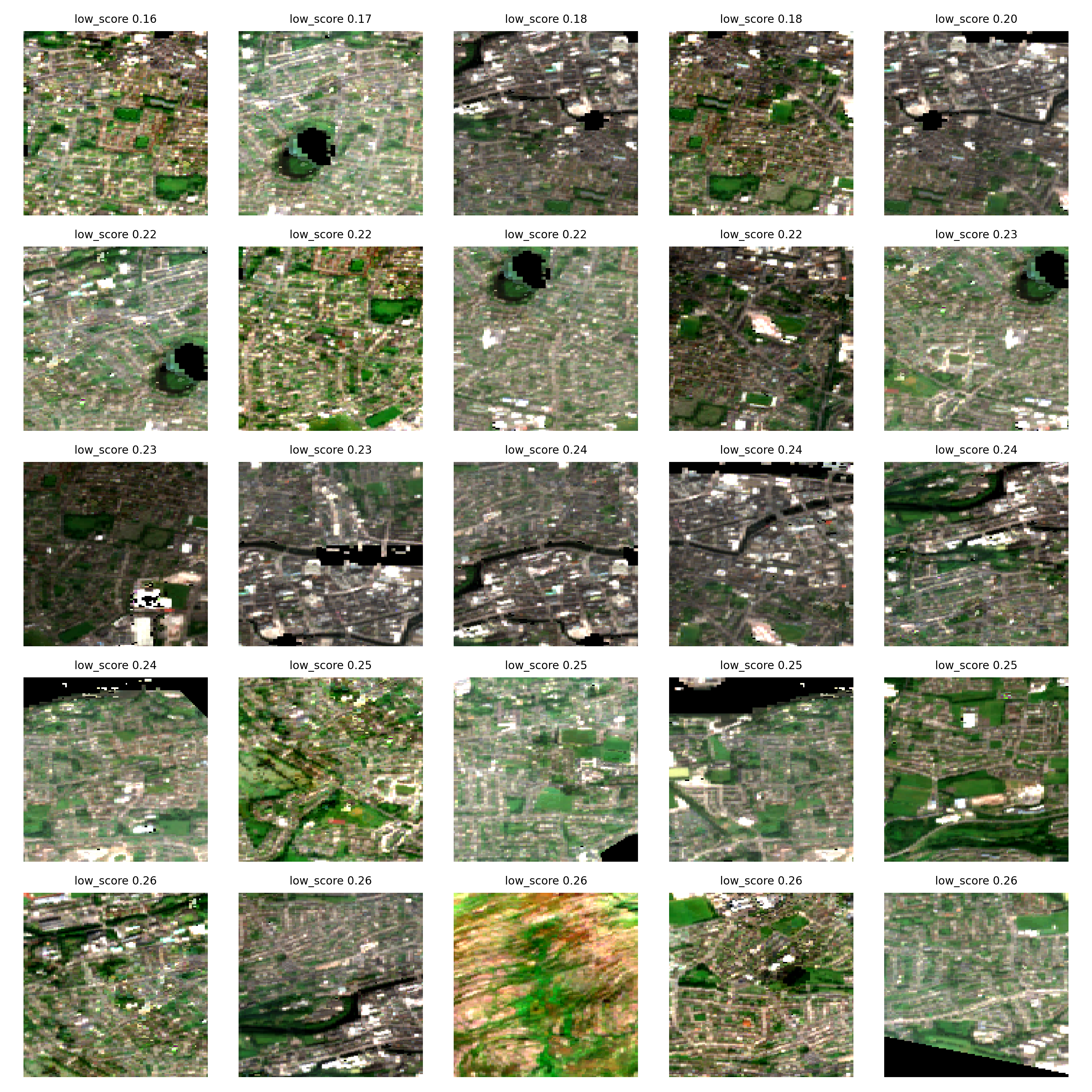}
        \centerline{Low score}
    \end{minipage}
    \caption{Qualitative examples from top and low score groups. Top score tiles tend to show pasture mosaics and farm-associated visual context, while low score tiles often show dense built areas, water, or other non-farm patterns. These examples support visual interpretation of the ranking signal, but they are not manual validation labels.}

    \label{fig:tile_examples}
\end{figure*}

{\color{black}
Figure~\ref{fig:cluster_overlay} overlays predicted clusters on the score field and held out OSM farm features. The clusters are distributed mainly across agricultural areas with strong pasture and greenness evidence. This pattern is consistent with the scoring design, which favors seasonal pasture, summer greenness, weak farm proximity, and spatial coherence.
\begin{figure}[t]
    \centering
    \includegraphics[width=0.92\linewidth]{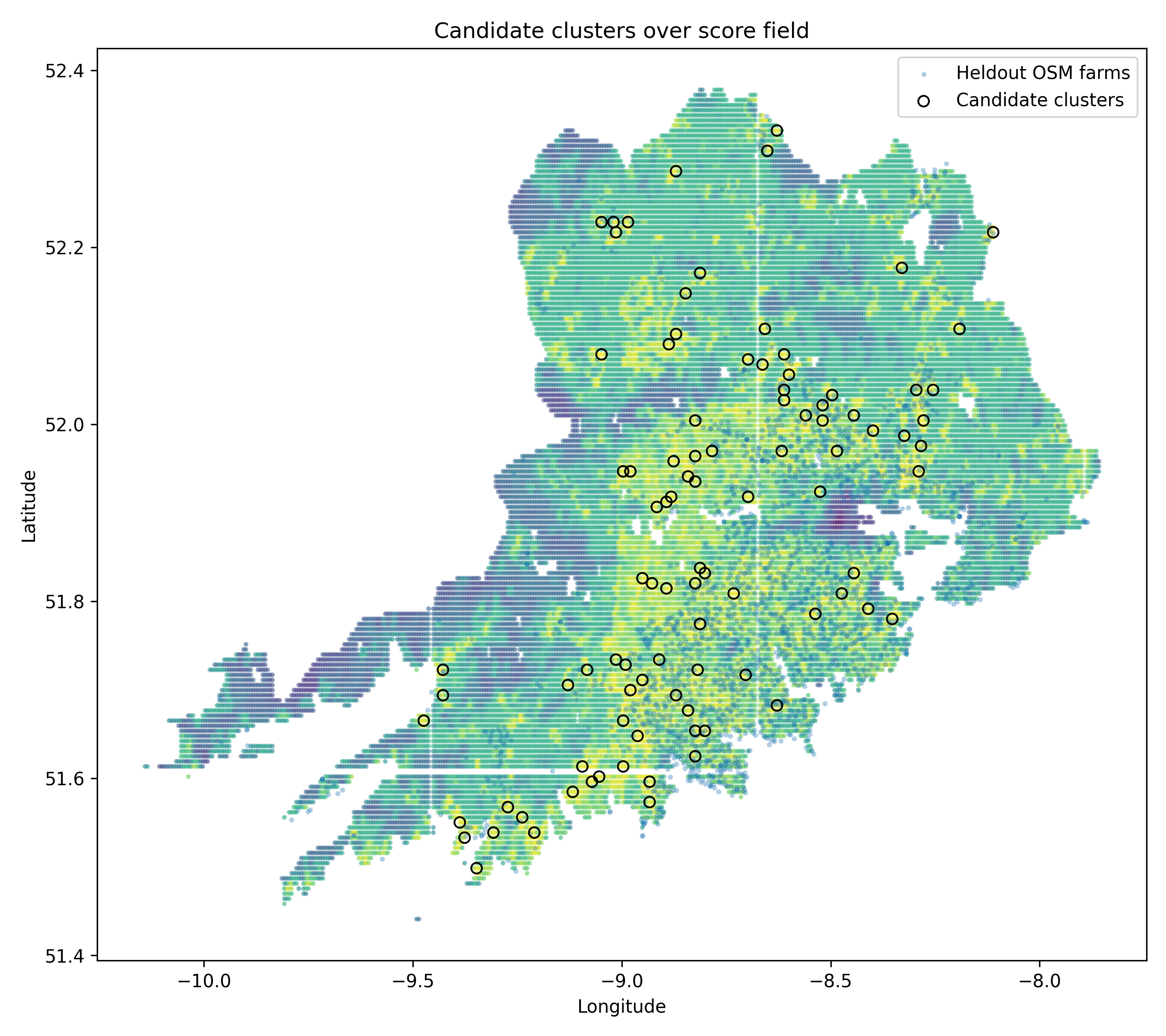}
    \caption{Candidate clusters over the score field and held out OSM farm features. Black circles indicate predicted candidate clusters. Held-out OSM features are used for proxy validation but not as farm proximity priors during scoring.}

    \label{fig:cluster_overlay}
\end{figure}

Figure~\ref{fig:tile_examples} shows qualitative tile examples from top and low score groups. Top-scoring tiles often contain pasture mosaics, field boundaries, rural roads, and farm-associated visual context. Low-score tiles often contain dense built-up areas, water, upland non-pasture areas, or other non-farm patterns. These examples support the interpretation of the ranking signal, but they are not used as manual validation.
}

\subsection{Ablation Study}
\label{subsec:ablation}
Table \ref{tab:ablation} compares scoring variants and baselines. Each variant uses the same 98th percentile selection rule and grid adjacency clustering. For direct comparison, clusters are ranked by their corresponding method score.

\begin{table*}[htbp]
\centering
{\color{black}
\caption{Ablation study using held out OSM farm features as proxy validation.}
\label{tab:ablation}
\begin{tabular}{lrrrrrr}
\hline
Method & Clusters & Median m & Top 5 500 m & Top 10 500 m & Top 10 1000 m & All 1000 m \\
\hline
Random score & 35 & 5263.2 & 0.20 & 0.20 & 0.40 & 0.17 \\
NDVI only & 113 & 5537.2 & 0.00 & 0.00 & 0.00 & 0.03 \\
Pasture only & 114 & 2912.3 & 0.20 & 0.10 & 0.10 & 0.29 \\
OSM prior only & 140 & 2652.2 & 0.00 & 0.00 & 0.10 & 0.23 \\
Rule only, no graph & 74 & 784.6 & 0.60 & 0.40 & 0.80 & 0.65 \\
No OSM farm prior & 124 & 3668.8 & 0.20 & 0.10 & 0.10 & 0.23 \\
Summer only & 87 & 5278.6 & 0.00 & 0.00 & 0.00 & 0.05 \\
\textbf{Full method} & \textbf{71} & \textbf{744.3} & \textbf{0.60} & \textbf{0.40} & \textbf{0.80} &\textbf{ 0.65} \\
\hline
\end{tabular}
}
\end{table*}

{\color{black}
The ablation study demonstrates that effective farm candidate ranking requires combining seasonal, geographic, and spatial-contextual signals. NDVI-only and summer-only variants perform poorly, indicating that single-season vegetation cues are insufficient for reliable candidate discovery. Pasture-only and OSM prior-only improve over these baselines but remain limited. The no-OSM farm prior variant also underperforms stronger variants, demonstrating the importance of weak geographic priors under incomplete supervision.

The rule-only variant performs well, producing 74 clusters with a median distance of 784.6 m, top-5 precision at 500 m of 0.60, top-10 precision at 500 m of 0.40, and top-10 precision at 1000 m of 0.80. The full method produces 71 clusters, reduces the median distance to 744.3 m, and maintains the same top-10 precision within 1000 m. This indicates that graph-based refinement preserves strong top-ranked precision while improving overall spatial consistency and reducing median distance to held-out OSM farm features.

}

\subsection{Score Diagnostics}
\begin{figure}[t]
    \centering
    \includegraphics[width=0.92\linewidth]{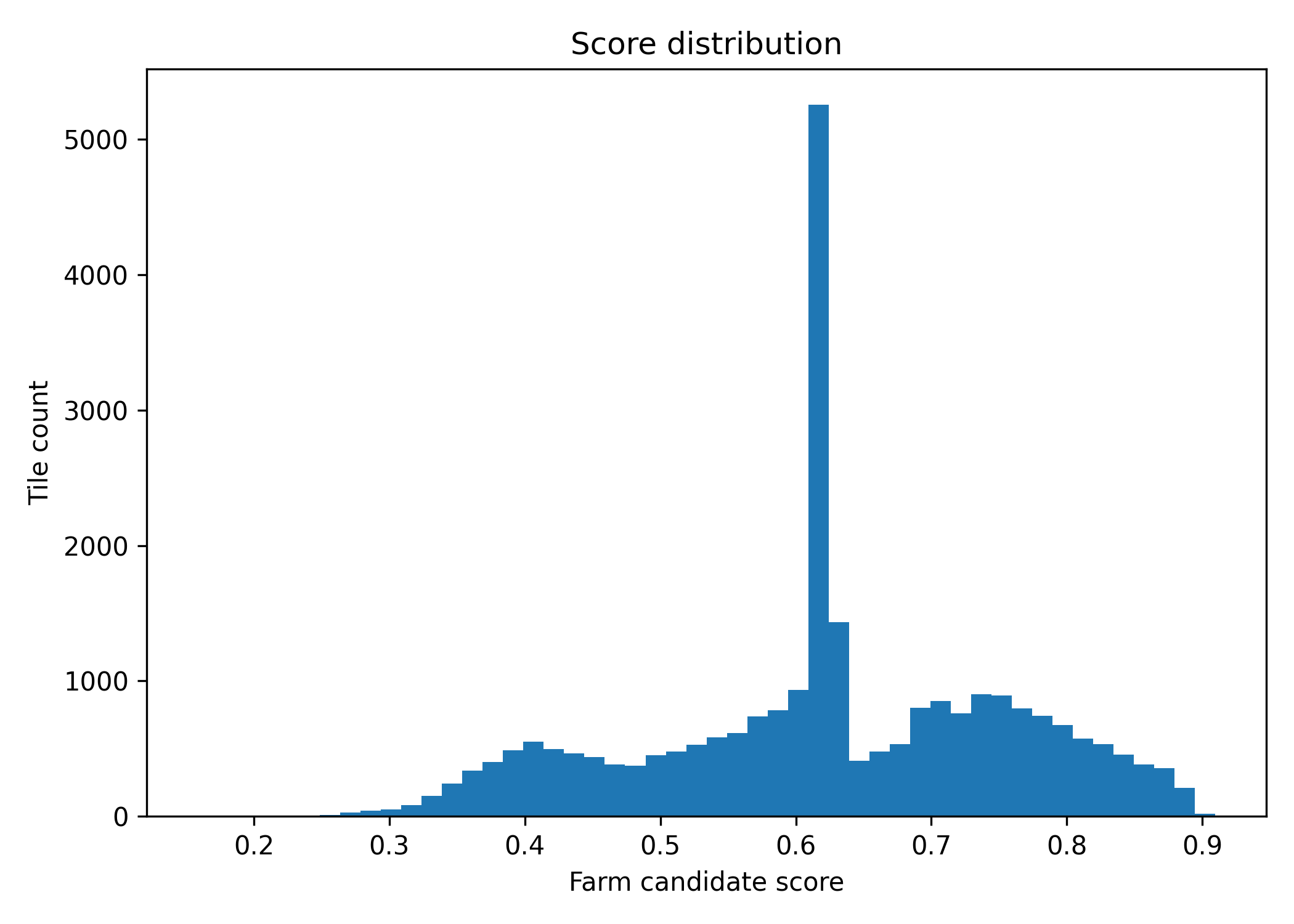}
    \caption{Distribution of tile scores. The clustering step selects the high-score tail. Scores are used for ranking and should not be interpreted as calibrated probabilities.}

    \label{fig:score_histogram}
\end{figure}

{\color{black}
Figure~\ref{fig:score_histogram} shows the distribution of tile scores. Most tiles fall below the high-confidence threshold, and candidate generation selects only the high-score tail. Since the score is a ranking heuristic, its absolute value should not be interpreted as a calibrated probability.
}

\section{Discussion and Conclusion}
\label{sec:discussion_conclusion}

{\color{black}
The experiments show that seasonal satellite imagery and weak geographic priors can support the discovery of farm candidates when direct farm labels are unavailable. The method reduces 26,722 multi-season tiles to 535 high-confidence tiles and 71 candidate clusters. In the main run, 46 of 71 clusters fall within 1000 m of held-out OSM farm features, and 80 percent of the top 10 candidates fall within 1000 m. These results indicate that the framework captures meaningful farm-associated spatial structure from seasonal satellite imagery under weak supervision.
}

\paragraph{Candidate discovery, not farm inventory.}
{\color{black}
The recall values are low because the method outputs a small number of high-confidence clusters compared with the number of held-out OSM farm features. This behavior is consistent with the framework's objective of candidate discovery. The objective is to prioritize likely farm-associated sites rather than recover every farm feature in the study area. Precision at rank is therefore more informative than total recall. The results should not be interpreted as a complete farm census. The results demonstrate that weak supervision can effectively reduce a large satellite-image collection into a compact ranked set of high-confidence agricultural candidates.
}


\paragraph{Role of self-supervised embeddings.}
The self-supervised encoder learns tile embeddings from multi-season imagery without direct labels. These embeddings are used for graph smoothing, allowing graph-based propagation across spatially adjacent tiles with similar seasonal structure. This is useful because farm evidence is spatially distributed across pasture, farmyard structure, roads, and field boundaries. The embeddings are not used as a standalone classifier. Their primary role is to support weakly supervised spatial reasoning and coherent candidate ranking.

\paragraph{Limits of proxy evaluation.}
Held out OpenStreetMap proximity is a proxy metric, not definitive ground truth. OSM farm features are incomplete, spatially uneven, and tag-dependent. A predicted cluster far from held-out OSM is not always a false positive, and a cluster near held-out OSM is not always a confirmed farm. The prior and held-out splits reduce direct reuse of the same farm features, but they do not eliminate all spatial dependence. Nearby OSM features may describe the same farm complex, and spatial dependence may remain even when the prior and held-out farm feature sets are disjoint. Future evaluation protocols could incorporate farm-complex-level splitting or spatial-block validation.

\paragraph{Localization and transfer.}
The use of $128 \times 128$ tiles provides useful spatial context but limits precise localization. A tile can contain pasture, roads, and farm structures distributed across multiple nearby locations, which helps explain why precision at 1000 m is stronger than at 500 m. Higher-resolution imagery, a smaller stride, or a second-stage localizer could improve localization accuracy. Although the method is not specific to Ireland, thresholds and feature weights may need recalibration in regions with different crop calendars, field sizes, settlement forms, or farm structures.

{\color{black}
Overall, this paper presents a weakly supervised pipeline for discovering dairy farm candidates from seasonal satellite imagery. The method combines multi-season Sentinel imagery, pasture and vegetation cues, weak OpenStreetMap farm priors, self-supervised tile embeddings, graph smoothing, and spatial clustering. Evaluation against held-out OpenStreetMap farm features shows that the main run achieves 0.60 top-5 precision within 500 m, 0.80 top-5 precision within 1000 m, and 0.80 top-10 precision within 1000 m. Since no manually verified farm ground truth is used, these results should be understood as proxy evidence for candidate ranking rather than definitive farm detection accuracy. More broadly, the results demonstrate the potential of weak supervision, self-supervised representation learning, and graph-based spatial reasoning for discovering agricultural candidates from large-scale satellite imagery.
}

\bibliographystyle{named}
\bibliography{ijcai26}

\end{document}